%% file: main.tex
\documentclass[letterpaper, 10 pt, conference]{ieeeconf}  

\usepackage{graphicx}
\usepackage{amsthm}
\usepackage{amsmath}
\usepackage{amssymb}
\usepackage{booktabs}
\usepackage{multirow}
\usepackage[accsupp]{axessibility}  %
\usepackage{graphicx}
\usepackage{amsmath}
\usepackage{amssymb}
\usepackage{booktabs}
\usepackage{tikz}
\usepackage{comment}
\usepackage{amsmath,amssymb} 
\usepackage{color}
\usepackage{indentfirst} 
\usepackage{float} 
\usepackage[accsupp]{axessibility} 
\usepackage{tabularx}
\usepackage{xspace}
\usepackage{amsfonts}
\usepackage{amssymb}
\usepackage{amsthm,amsmath}
\usepackage{mathrsfs}
\usepackage{indentfirst}
\usepackage{multirow}

\IEEEoverridecommandlockouts         
\definecolor{myblue}{rgb}{.39,.58,.93}

\title{\LARGE \bf
CrossVideo: Self-supervised Cross-modal Contrastive Learning \\for Point Cloud Video Understanding
}

\author{Yunze Liu$^{1,2}$, Changxi Chen$^{1}$, Zifan Wang$^{1}$, Li Yi$^{1,2,3}$
\thanks{*This work was supported by Shanghai Qi Zhi Institute}
\thanks{$^{1}$Tsinghua University, $^{2}$ Shanghai Qi Zhi Institute, $^{3}$ Shanghai Artificial Intelligence Laboratory.
        {\tt\small Mails:liuyzchina@gmail.com}}%
}

\begin{document}

\maketitle
\thispagestyle{empty}
\pagestyle{empty}

\begin{abstract}


This paper introduces a novel approach named CrossVideo, which aims to enhance self-supervised cross-modal contrastive learning in the field of point cloud video understanding. Traditional supervised learning methods encounter limitations due to data scarcity and challenges in label acquisition. To address these issues, we propose a self-supervised learning method that leverages the cross-modal relationship between point cloud videos and image videos to acquire meaningful feature representations. Intra-modal and cross-modal contrastive learning techniques are employed to facilitate effective comprehension of point cloud video. We also propose a multi-level contrastive approach for both modalities. Through extensive experiments, we demonstrate that our method significantly surpasses previous state-of-the-art approaches, and we conduct comprehensive ablation studies to validate the effectiveness of our proposed designs.

\end{abstract}

\section{INTRODUCTION}

\input{tex/intro}

\section{RELATED WORK}

\input{tex/related}

\section{METHOD}

\input{tex/method}

\section{EXPERIMENT}

\input{tex/exp}

\section{CONCLUSION}

\input{tex/conclusion}







\bibliography{references}
\bibliographystyle{IEEEtran}

\end{document}

%% file: tex/intro.tex
\pdfoptionpdfminorversion = 7
Recently, there has been a wide interest in understanding point cloud sequences in 4D (3D space + 1D time)~\cite{fan21p4transformer,wen2022point,shi2022weakly,khurana2023point}.
Point cloud video is a dynamic sequence constructed based on three-dimensional point cloud data. It plays an important role in the fields of computer vision and machine learning. Point cloud video contains rich geometric and topological information, which can accurately describe objects and scenes in the real world. It has wide applications in areas such as autonomous driving, robot navigation, and augmented reality. 

Despite the ubiquity of 4D data, annotating such data on a large scale with detailed information is costly. Therefore, we need to find ways to make use of massive amounts of unlabeled data.
Among possible solutions, self-supervised representation learning has demonstrated its effectiveness in various fields, including image, video, and point cloud.

The existing self-supervised point cloud video representation learning solution is 4D distillation~\cite{zhang2023complete}. The core of this method is to compare the differences between different frames of the same object or scene, which can capture features such as shape, size, and dynamic changes of objects. A teacher network that can obtain full information is used to guide a student network that can only obtain partial information. However, real point cloud videos are often hindered by sampling patterns, noise, and other interferences, which often prevent the student network from recovering full information. Besides, this method also requires manually designing inputs for both the teacher network and the student network, which significantly affects pretraining performance.


We assume the appearance information in image videos can complement point cloud videos. For example, when a person is ``graffitiing on a wall'', it is difficult to perceive the changes on the wall from the point cloud alone, but images can provide information about the changes.
Additionally, the color information and texture details in image video can also provide additional features to assist in motion and action recognition. We assume that point cloud video and image video can help each other, so we propose to learn 4D representation by leveraging the synergy from image video.

\begin{figure}[t]
    \centering
    \includegraphics[width=1\linewidth]{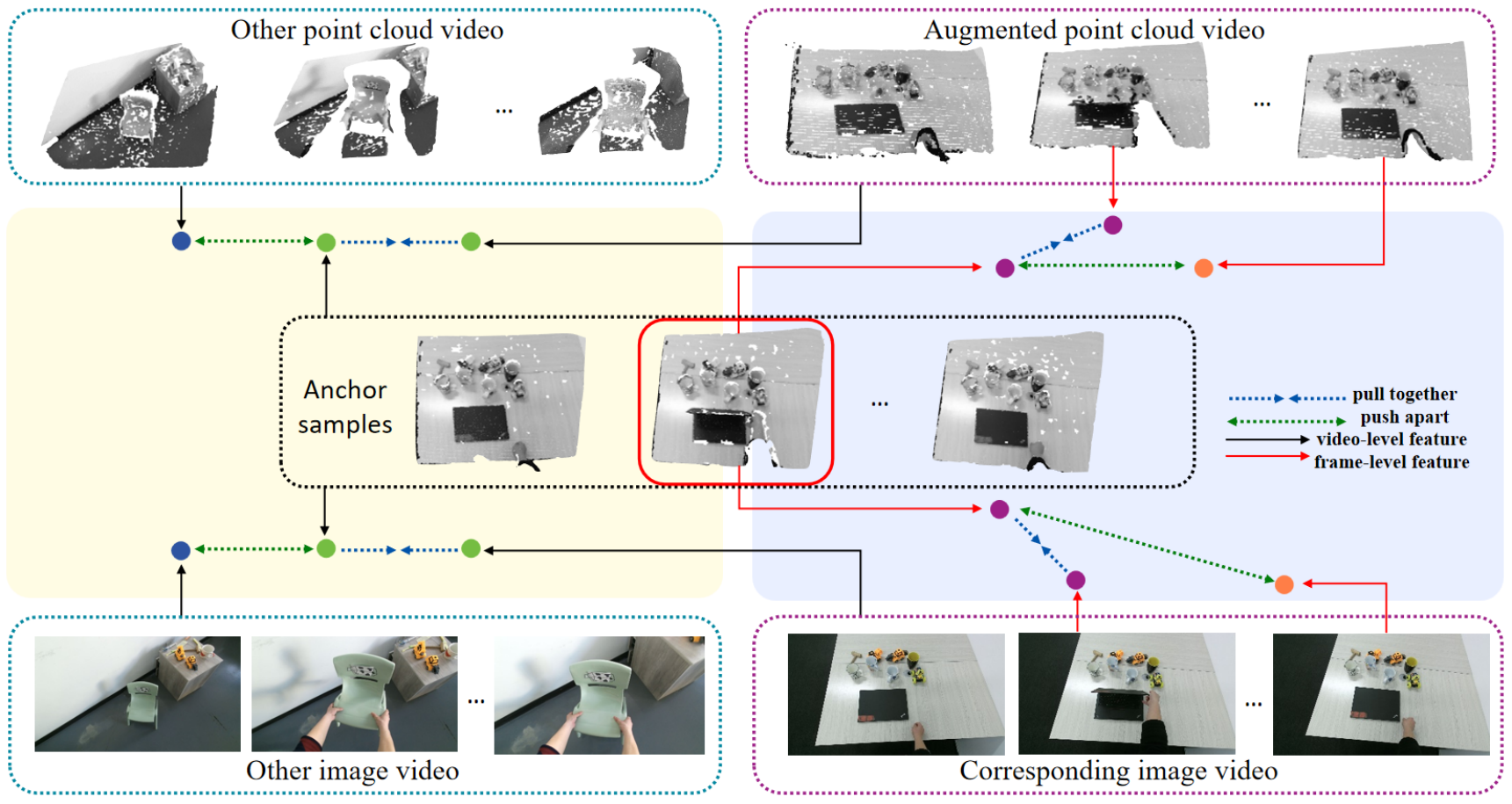}
    \caption{Our core idea is to learn 4D point cloud video representations by leveraging the synergy from image video under a self-supervised manner. We propose intra-modal contrastive objective and cross-modal contrastive objective to learn the spatio-temporal invariance of point cloud videos and the correlation between point cloud videos and image videos.}
    \label{fig:teaser}
    \vspace{-5mm}
\end{figure}

But learning 4D representations by leveraging the synergy from image video under a self-supervised manner is not an easy task. It is well known that a point cloud video is composed of multiple point clouds, and the same applies to image videos. We need to enable the network to understand information at different granularities, including understanding at the video level and at the video frame level. Besides, to conduct effective contrastive learning, it is also necessary to design a suitable backbone for cross-modal pre-training. 


We have studied the characteristics of point cloud videos and image videos, and have obtained two main observations. Firstly, for point cloud videos, as they provide accurate three-dimensional coordinate information, they are able to capture the spatial position, shape, and motion trajectory of objects. 
Appearance information in image videos can complement point cloud videos, like the example of ``graffiti on a wall'' mentioned above. Continuous sequences of images in image videos can reflect the motion patterns and action states of objects from color information and texture details. 
Therefore, these two modalities naturally complement each other. To facilitate the synergy between image videos and point cloud videos, we introduce a self-supervised cross-modal contrastive learning method.
Secondly, there exists a strong correlation between video and frames, thus understanding videos can enhance the understanding of frames, and vice versa. Therefore, we need to design objective to reflect such synergy by contrasting features at different granularity levels.

Cross-modal contrastive learning for point cloud video understanding has been rarely discussed in the self-supervised representation learning literature. 
To the best of our knowledge, we are the first to propose to learn 4D representations by leveraging the synergy from image video under a self-supervised manner. We have carefully designed a multi-level objective function and pre-training backbone. 

In particular, our method is as shown in Fig.~\ref{fig:teaser}. 
First, we design a network to extract video-level and frame-level features from point cloud videos and image videos. In addition, for point cloud videos, we apply spatio-temporal data augmentation to obtain a new version.
Then, we propose intra-modal contrastive objective and cross-modal contrastive objective to learn the spatio-temporal invariance of point cloud videos and the correlation between point cloud videos and image videos.
Finally, we can get a strong point cloud video encoder by self-supervised pretraining.

We evaluate our method on two downstream point cloud video tasks: 4D action segmentation on HOI4D~\cite{Liu_2022_CVPR}, 4D semantic segmentation on HOI4D~\cite{Liu_2022_CVPR}. We demonstrate significant improvements over
the previous method($+2.6\%$ accuracy on HOI4D action segmentation, $+1.5\%$ mIoU on HOI4D semantic segmentation). At the same time, we also demonstrate that the image video encoder can also integrate information from the point cloud video, thereby enhancing its representation capability.

The contributions of this paper are fourfold: First, we propose the first 4D self-supervised cross-modal representation learning method which facilitates the synergy of image video and point cloud video learning. Second, we propose to use intra-modal and cross-modal contrastive learning to facilitate an effective point cloud video understanding. Third, we propose to contrast the features of both modalities at different levels. Fourth, extensive experiments show that our method outperforms previous state-of-the-art methods by a large margin and we provide comprehensive ablation studies to validate our designs.

%% file: tex/related.tex
\subsection{4D Point cloud video understanding.}
Compared to understanding static 3D point clouds, comprehending 4D point cloud sequences requires a stronger focus on aggregating and utilizing spatial-temporal information to perceive both the geometry and dynamics.
To tackle these challenges, several 4D backbones have been proposed, which can be divided into two categories based on their representations. The first category involves voxelizing raw point clouds and extracting features from 4D voxels. An example is MinkowskiNet \cite{choy20194d}, which applies 4D spatial-temporal convolutions on 4D voxels.
The second category operates directly on raw points. For instance, MeteorNet \cite{liu2019meteornet} extends PointNet++ \cite{qi2017pointnet++} by introducing a temporal dimension and explicitly tracking points' motion for grouping. PSTNet \cite{fan2022pstnet}, on the other hand, constructs a point tube along the temporal dimension for 4D point convolution.
State-of-the-art methods like Point 4D Transformer \cite{fan21p4transformer} and PPTr \cite{wen2022point} belong to the second category. They incorporate transformer architecture to avoid point tracking and better capture spatio-temporal correlation.

\subsection{3D representation learning}
Due to advances in 2D representation learning~\cite{he2020momentum,he2022masked,liu2021contrastive}, similar progress has been made in the field of 3D representation learning~\cite{pang2022masked,yu2021pointbert,zhang2022pointclip,liu2020p4contrast}.
Existing methods fall into two main categories: generative-based methods~\cite{bear2020learning,wang2021unsupervised,yu2021pointbert,zhang2022point}, such as Point-Bert~\cite{yu2021pointbert} which recover masked object parts to learn Transformer representations, and context-based methods~\cite{xie2020pointcontrast,rao2021randomrooms,zhang2021self,chen2021shape,hou2021exploring}, like SelfCorrection~\cite{chen2021shape} which distinguish and restore destroyed objects to learn informative representations.
Existing methods face challenges when extended to high-dimensional 4D data due to the difficulty of optimization and computational cost. 
We aim to explore the extraction of high-quality spatio-temporal features at the scene level for enhancing 4D downstream tasks, so we introduce the utilization of multimodal data into the field of 4D point cloud pretraining.



\subsection{4D Representation Learning.}
4D Representation Learning is an emerging field upon 3D research. For instance, 4Dcontrast \cite{chen20224dcontrast} exploits 4D motion information and STRL \cite{huang2021spatio} uses temporal-spatial contrastive learning. Although pre-trained on 4D data, they only learn static 3D representations. C2P \cite{zhang2023complete} is a new pre-training method for 4D point cloud sequence
representation learning. Our approach builds upon the foundations laid by STRL and C2P,  which represents a pioneering effort in incorporating multimodal data into 4D point cloud pretraining.


\subsection{Cross-modal Representation Learning.}

Cross-modal learning leverages diverse data sources to yield rich contextual information and effective semantic comprehension. 
Recent studies \cite{arandjelovic2017look,desai2021virtex,radford2021learning} have demonstrated the efficiency of cross-modal pretraining in generating various representations suitable for many subsequent tasks.
CLIP \cite{radford2021learning} learns a multimodal embedding space by maximizing cosine similarity between images and text, while Morgado et al. \cite{morgado2021audio} combines audio and video modalities to achieve notable performance improvements in action and sound recognition tasks through cross-modal agreement. 

In addition, the research conducted by  \cite{xu2021image2point} expands the pretrained 2D image model to a point-cloud model by utilizing filter inflation. \cite{Afham_2022_CVPR} suggest rendering images for each point cloud and implementing cross-modal contrastive loss to enhance performance. However, these approaches cannot be directly applied to 4D point cloud videos due to the unsuitability and ineffectiveness of existing pre-training backbones and contrastive objectives.


%% file: tex/method.tex
\pdfoptionpdfminorversion = 7
\subsection{Overview}
We propose a novel multimodal point cloud video pretraining method to obtain a powerful point cloud video encoder in an unsupervised manner. Our key idea is to learn 4D representations by leveraging the synergy from image video. And we propose to align point cloud videos with image videos at both the video-level and frame-level in the feature space through cross-modal contrastive learning. We note that since our main focus is on the understanding of point cloud videos, we will primarily introduce how to train a powerful point cloud video encoder in the following sections. At the same time, in the experimental section, we also demonstrate the 2D video encoder can also obtain strong representations.

\begin{figure}[t]
    \centering
    \includegraphics[width=1\linewidth]{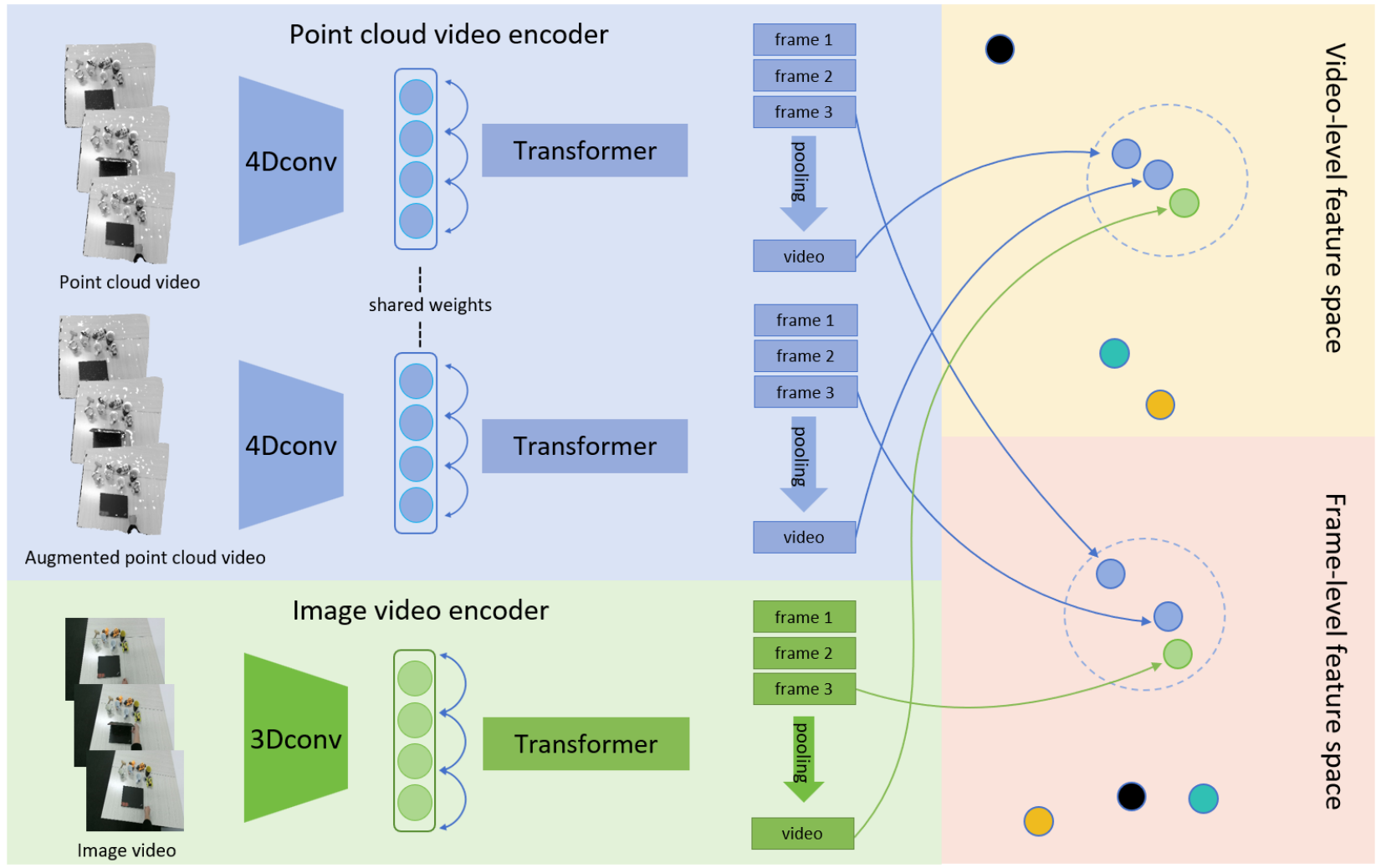}
    \caption{Cross-modality pretraining enables distillation of knowledge from 2D videos into point cloud videos, while intra-modality pretraining allows the encoder to possess invariance to a set of geometric transformations. The feature space at the video level and the frame level can both provide constraints on the encoder at different levels of granularity.}
    \label{fig:method}
    \vspace{-2mm}
\end{figure}

The overview of our method is shown in the Fig.~\ref{fig:method}. Given a point cloud video, we first perform data augmentation to obtain an augmented version of the point cloud video. A 4Dconv module is used to extract features, followed by a transformer to further facilitate feature communication across different time steps. Consequently, we can obtain the features of the point cloud video and its features for each frame. Similarly, for the image video, we first use a 3Dconv module to extract features and a transformer to facilitate feature communication, obtaining the features of the image video and its features for each frame. At the video-level and frame-level, we perform cross-modal and intra-modal contrastive learning respectively to obtain powerful encoders for point cloud video and image video. Cross-modal pre-training allows knowledge to be transferred from image video to the point cloud video encoder, while intra-modal pre-training enables the encoder to have invariance to a set of geometric transformations. The feature spaces at both the video-level and frame-level provide constraints on the encoder at different granularities, thereby further enhancing its representation ability. In the remaining part of this section, we will introduce our method in detail. First, we present the network architecture details of the proposed method. Then, we describe the contrastive learning loss functions formulated under intra-modal and cross-modal settings. Finally, we propose our overall training objective.


\subsection{Preliminaries}

Suppose we are given a dataset, $\mathcal{D} = \{\left(\textbf{P}_i, \textbf{I}_i\right)\}_{i=1}^{\vert\mathcal{D}\vert}$ with $\textbf{P}_i \in \mathbb{R}^{L \times N \times 3}$ and $\textbf{I}_i \in \mathbb{R}^{L \times H \times W \times 3}$. 
We define each $\textbf{P}_i$ with sequence length L as $\textbf{P}_i = \{s_1, s_2, \dots,s_j,\dots, s_L\}$ where each $s$ denotes one frame. 
We aim to train a point cloud video feature extractor $f_{\theta_\textbf{P}}\left(.\right)$ in a self-supervised manner to be effectively transferable to downstream tasks. To this end, we use an image video feature extractor $f_{\theta_\textbf{I}}\left(.\right)$, multi-layer perceptron (MLP) projection heads $g_{\phi_\textbf{P}}\left(.\right)$ and $g_{\phi_\textbf{I}}\left(.\right)$ for point cloud and image respectively.

\subsection{Pre-training Backbone}
For point cloud videos, due to providing accurate 3D coordinate information, they can capture the spatial position, shape, and motion trajectory of objects. This advantage makes point cloud videos beneficial for motion analysis and action recognition. 

Although image video cannot directly provide three-dimensional geometric information, the captured continuous image sequences can reflect the motion patterns and action states of objects. By utilizing the temporal relationship of image sequences, motion analysis and action recognition can be performed. Additionally, the color information and texture details in image video can also provide additional features to assist in motion and action recognition.

In order to support our pre-training, we need a network that can extract features from both image videos and point cloud videos and can support two types of features(video-level and frame-level). Based on this, we first design a three-tower structure as shown in Fig~\ref{fig:method}. 
The whole network is divided into two branches, namely the point cloud-video branch and the image-video branch.

\subsubsection{Point cloud video encoder}
The point cloud video branch consists of two shared-weight networks, each taking the original point cloud video and the enhanced point cloud video as inputs.
Point cloud video encoder is divided into 4D convolution module and Transformer module. The 4D convolution is mainly used to extract spatiotemporal features, while the Transformer is used to enhance information exchange across different time sequences.
\subsubsection{Image video encoder}

Image video encoder is based on Asformer, which is mainly divided into 3D Convolution and Transformer modules. 3D Convolution is used to extract spatiotemporal features, while Transformer is used to enhance the model's representation ability and increase cross-temporal communication capability.

\subsection{Intra-Modal Contrastive learning}
Inspired by contrastive learning in the fields of image and static point cloud, we believe that intra-modal contrastive learning is crucial for encoding invariant representations. 
At the video level, we model the problem to keep invariance to spatiotemporal data augmentation including geometric transformations $\mathbf{T_g}$ and temporal transformations $\mathbf{T_t}$. 

Given an input 4D point cloud video $\textbf{P}_i$, we denote the raw point cloud video and the augmented versions $\textbf{P}_i^{t_1}$ and $\textbf{P}_i^{t_2}$ of it. We compose $t_2$ by randomly combining transformations from $\mathbf{T_g}$ and $\mathbf{T_t}$. For geometric transformations $\mathbf{T_g}$, we use rotation, scaling and translation. For temporal transformation, we extract some frames by down-sampling. 

The point cloud video feature extractor $f_{\theta_\textbf{P}}$ take $\textbf{P}_i$ as input, and output both video-level and frame-level feature. 
The feature vectors are projected to an invariant space $\mathbb{R}^d$ where the contrastive loss is applied, using the projection head $g_{\phi_\textbf{P}}$. 
We denote the projected video-level feature of $\textbf{P}_i^{t_1}$ and $\textbf{P}_i^{t_2}$ as $\textbf{z}_{v_i}^{t_1}$ and $\textbf{z}_{v_i}^{t_2}$ and projected frame-level feature as $\textbf{z}_{f_i}^{t_1}$ and $\textbf{z}_{f_i}^{t_2}$ respectively where, $\textbf{z}_i^{t} = g_{\phi_\textbf{P}}\left(f_{\theta_\textbf{P}}\left(\textbf{P}_i^{t}\right)\right)$. Here, $\textbf{z}_{v_i}^{t}$ is obtained by applying the max-pooling to $\textbf{z}_{f_i}^{t}$.

For video-level contrastive learning, the goal is to maximize the similarity of $\textbf{z}_{v_i}^{t_1}$ with $\textbf{z}_{v_i}^{t_2}$ while minimizing the similarity with all the other projected vectors in the mini-batch of point clouds. Similarly, for frame-level contrastive learning, the goal is to maximize the similarity of $\textbf{z}_{f_i}^{t_1}$ with $\textbf{z}_{f_i}^{t_2}$ while minimizing the similarity with others.   
We leverage NT-Xent loss proposed in SimCLR for instance discrimination. 

For video-level, We compute the loss function $L_v(i,{t_1},{t_2})$ for the positive pair of examples $\textbf{z}_{v_i}^{t_1}$ and $\textbf{z}_{v_i}^{t_2}$ as:

\begin{equation}
\footnotesize
\label{eq:l1()}
    L_v = -\log \frac{\exp(s(\textbf{z}_{v_i}^{t_1}, \textbf{z}_{v_i}^{t_2})/\tau)}{ \sum\limits_{\substack{k=1 \\ k \neq i}}^{N} \exp(s(\textbf{z}_{v_i}^{t_1}, \textbf{z}_{v_k}^{t_1})/\tau) + \sum\limits_{\substack{k=1}}^{N} \exp(s(\textbf{z}_{v_i}^{t_1}, \textbf{z}_{v_k}^{t_2})/\tau)}
\end{equation}
where N is the mini-batch size, $\tau$ is the temperature co-efficient and $s\left(.\right)$ denotes the cosine similarity function. 

Similarly, for frame-level, the objective can be written as: 
\begin{equation}
\footnotesize
\label{eq:l2()}
    L_f = -\log \frac{\exp(s(\textbf{z}_{f_i}^{t_1}, \textbf{z}_{f_i}^{t_2})/\tau)}{ \sum\limits_{\substack{k=1 \\ k \neq i}}^{N} \exp(s(\textbf{z}_{f_i}^{t_1}, \textbf{z}_{f_k}^{t_1})/\tau) + \sum\limits_{\substack{k=1}}^{N} \exp(s(\textbf{z}_{f_i}^{t_1}, \textbf{z}_{f_k}^{t_2})/\tau)}
\end{equation}

Our intra-modal instance discrimination loss function $\mathcal{L}_{intra}$ for a mini-batch can be described as:

\begin{equation}
\label{eq:imid}
    \mathcal{L}_{intra} = \frac{1}{2N} \sum_{i=1}^{N}[L_v(i,{t_1},{t_2}) + L_f(i,{t_1},{t_2})]
\end{equation}

\subsection{Cross-Modal Contrastive learning}


We designed a cross-modal pre-training method to empower the encoders of both modalities, facilitating communication and integration between the two modalities.
To the best of our knowledge, we are the first study to systematically investigate the pretraining of multi-modal point cloud videos. And we propose a simple and effective method.  We empirically validate with the experimental results that our objective outperforms existing unsupervised representation methods, thus facilitating an effective representation learning of 4D point clouds video.

To this end, we first embed the corresponding image video $\textbf{I}_i$ to a feature space using the image video feature extractor $f_{\theta_\textbf{I}}$.  We then project the feature vectors to the invariant space $\mathbb{R}^d$ using the image projection head $g_{\phi_\textbf{I}}$. Note that, we can obtain both video-level features and frame-level features here. The projected image video-level and frame-level feature is defined as $\textbf{h}_{v_i}$ and $\textbf{h}_{f_i}$ where $\textbf{h}_i =  g_{\phi_\textbf{I}}\left(f_{\theta_\textbf{I}}\left(\textbf{I}_i\right)\right)$. Here, $\textbf{h}_{v_i}$ is obtained by applying the max-pooling to $\textbf{h}_{f_i}$.
Then, we compute the mean of the projected vectors $\textbf{z}_{v_i}^{t_1}$ and $\textbf{z}_{v_i}^{t_2}$ to obtain the projected prototype video-level feature $\textbf{z}_{v_i}$ of $\textbf{P}_i$. Similarly, we can also obtain frame-level features $\textbf{z}_{f_i}$ of $\textbf{P}_i$.

\begin{equation}
    \textbf{z}_{v_i} = \frac{1}{2}\left(\textbf{z}_{v_i}^{t_1} + \textbf{z}_{v_i}^{t_2} \right);      
    \textbf{z}_{f_i} = \frac{1}{2}\left(\textbf{z}_{f_i}^{t_1} + \textbf{z}_{f_i}^{t_2} \right)
\end{equation}

In the invariance space, we aim to maximize the similarity of $\textbf{z}_{v_i}$ and $\textbf{z}_{f_i}$ with $\textbf{h}_{v_i}$ and $\textbf{h}_{f_i}$, respectively. 
We compute the loss function $l(i, \textbf{z},\textbf{h})$ for the positive pair of examples as:

\begin{equation}
\footnotesize
\label{eq:c1()} 
    C_v = -\log \frac{\exp(s(\textbf{z}_{v_i}, \textbf{h}_{v_i})/\tau)}{ \sum\limits_{\substack{k=1 \\ k \neq i}}^{N} \exp(s(\textbf{z}_{v_i}, \textbf{z}_{v_k})/\tau) + \sum\limits_{\substack{k=1}}^{N} \exp(s(\textbf{z}_{v_i}, \textbf{h}_{v_k})/\tau)}
\end{equation}
\begin{equation}
\footnotesize
\label{eq:c2()}
    C_f = -\log \frac{\exp(s(\textbf{z}_{f_i}, \textbf{h}_{f_i})/\tau)}{ \sum\limits_{\substack{k=1 \\ k \neq i}}^{N} \exp(s(\textbf{z}_{f_i}, \textbf{z}_{f_k})/\tau) + \sum\limits_{\substack{k=1}}^{N} \exp(s(\textbf{z}_{f_i}, \textbf{h}_{f_k})/\tau)}
\end{equation}

where s, N, $\tau$ refers to the same parameters as in Eq. \ref{eq:l1()}. The cross-modal loss function $\mathcal{L}_{cmid}$ for a mini-batch is then formulated as:

\begin{equation}
\label{eq:cmid}
    \mathcal{L}_{cross} = \frac{1}{2N} \sum_{i=1}^{N}[C_v(i, \textbf{z}, \textbf{h}) + C_f(i, \textbf{z}, \textbf{h})]
\end{equation}

\subsection{Overall Contrastive Objective}

Finally, we obtain the resultant loss function during training as the combination of $\mathcal{L}_{imid}$ and $\mathcal{L}_{cmid}$ where $\mathcal{L}_{imid}$ imposes invariance to point cloud transformation while $\mathcal{L}_{cmid}$ injects the 3D-2D correspondence.

\begin{equation}
    \mathcal{L}_{total} = \mathcal{L}_{intra} + \mathcal{L}_{cross}
\end{equation}

%% file: tex/exp.tex
In this section, following other representation learning methods, we use the pre-trained network weights as initialization and fine-tune them in 4D downstream tasks. The performance gain will be a good indicator of measuring the quality of the learned feature. In this section, we cover two 4D point cloud sequence understanding tasks: action segmentation on HOI4D, semantic segmentation on HOI4D in Section~\ref{sec:HOI4Das}, \ref{sec:HOI4Dss} respectively. For these tasks, some basic pretraining settings will first be introduced in Section~\ref{sec:pretrain}. We demonstrate that the 2D video encoder involved in pre-training can also obtain strong representations in Section~\ref{sec:2d}.
We also provide extensive ablation studies to validate our design choices in Section~\ref{sec:ablation}.

\subsection{Pre-training}
\label{sec:pretrain}
We use HOI4D\cite{Liu_2022_CVPR} as the dataset for pretraining. HOI4D consists of 2.4M
RGB-D egocentric video frames over 4000 sequences interacting with 800 different object instances from 16 categories over 610 different indoor rooms.
For a given RGB-D sequences, we first project each depth map to the point cloud. We also use the corresponding RGB images for pretraining. We use 2048 points for each point cloud while we resize the corresponding RGB image to $224 \times 224$. In addition to the augmentations applied for point cloud as described in Sec, we perform random crop and color jittering for rendered images as data augmentation.       

To draw a fair comparison with the existing methods, we deploy PPTr\cite{wen2022point} and P4Transformer\cite{fan21p4transformer} as the point cloud feature extractors.  Our experiments show that our method consistently outperforms previous approaches in both feature extractors. 
We use a 3Dconv and Transformer module as the image video feature extractor. We employ a 2-layer MLP as the projection heads which yields a 256-dimensional feature vector projected in the invariant space $\mathbb{R}^d.$ 

We use SGD optimizer to train the network. The learning rate is set to be 0.01 and we use a learning rate warmup for 5 epochs, where the learning rate increases linearly for the first 10 epochs. 
The dimension of the features used to calculate the contrastive loss is set to 2048. The temperature coefficient used when calculating contrastive loss is set to 0.07. The time window size is set to 3. 
As for P4DConv, we set the spatial stride to 32, the radius of the ball query is set to 0.9 and the number of samples is set to 32 by default. With batch-size set to 8, our pre-training method can be implemented on a NVIDIA GeForce 3090 GPUs.
After pre-training we discard the image feature extractor $f_{\theta_\textbf{I}}\left(.\right)$ and projection heads $g_{\phi_\textbf{P}}\left(.\right)$ and  $g_{\phi_\textbf{I}}\left(.\right)$. All downstream tasks are performed on the pre-trained point cloud feature extractor $f_{\theta_\textbf{P}}\left(.\right)$.

\subsection{Fine-tuning on HOI4D 4D action segmentation}
\label{sec:HOI4Das}
\textbf{Setup.} To demonstrate the effect of our approach, we conduct experiments on the HOI4D action segmentation task. For each point cloud sequence, we need to predict the action label for each frame. We follow the official data split of HOI4D with 2971 training scenes and 892 test scenes. Each sequence has 150 frames, and each frame has 2048 points.  
We also conduct experiments with other 4D pre-training strategies including STRL(a 4D spatial-temporal contrastive learning method)\cite{huang2021spatio} and VideoMAE(an MAE-based method for video representation learning)\cite{tong2022videomae} and C2P(a 4D Distillation method)\cite{zhang2023complete}. The following metrics are reported: framewise accuracy (Acc), segmental edit distance, as well as segmental F1 scores at the overlapping thresholds of 10\%, 25\%, and 50\%. Overlapping thresholds are determined by the IoU ratio. 

\textbf{Result.} As reported in Tab.~\ref{table:hoi4das}, our method has big improvement on both two backbones. For the state-of-the-art backbone PPTr, it can be seen that our method consistently outperforms STRL and VideoMAE by a big margin for all metrics. Considering STRL, its short sequence augmentation can not well guide the model to notice motion cues so it is very hard to leverage temporal information which is actually very important in action segmentation tasks. Simple extension of VideoMAE to point cloud sequence also shows very little improvement. Unlike video pixel tokens which are regular and compact, high down-sampled point cloud tokens have very irregular and sparse patterns. It is hard for the model to learn to predict raw points. Also notice that VideoMAE do self-supervised learning on a point level, which is very hard to learn motion features.
C2P formulate 4D self-supervised representation learning as a teacher-student knowledge distillation framework and let the student learn useful 4D representations with the guidance of the teacher. 
However, it requires complex and time-consuming data generation steps to generate a partial observed point cloud video as a form of data augmentation. This significantly increases the computational cost during pre-training and relies on continuous camera viewpoint generation.
Our method is introduced to leverage image video to help the understanding of point cloud videos which finally comes in a satisfying result. 

\begin{table}
\setlength{\tabcolsep}{0.15mm}
\scriptsize{
\begin{center}
\caption{4D Action segmentation on HOI4D dataset}
\label{table:hoi4das}
\newcolumntype{Y}{>{\centering\arraybackslash}X}
{
\begin{tabularx}{\columnwidth}{>{\centering} m{0.3\columnwidth}|>{\centering} m{0.12\columnwidth}|>{\centering} m{0.125\columnwidth}|Y|Y|Y|Y}
\hline\noalign{\smallskip}
Method & Frames   & Acc & Edit & F1@10 & F1@25 & F1@50\\
\noalign{\smallskip}
\hline
\noalign{\smallskip}

P4Transformer & 150 & 71.2 & 73.1 & 73.8 & 69.2& 58.2\\
P4Transformer+C2P & 150 & 73.5 & 76.8 & 77.2 & 72.9 & 62.4\\
P4Transformer+ours & 150 & \textbf{76.2} & \textbf{79.2} & \textbf{78.8} & \textbf{75.4} & \textbf{65.0}\\
\noalign{\smallskip}
\hline
\noalign{\smallskip}
PPTr & 150 &  77.4 & 80.1 & 81.7 & 78.5 & 69.5\\
PPTr+STRL & 150 & 78.4 & 79.1 & 81.8 & 78.6 & 69.7 \\
PPTr+VideoMAE & 150 & 78.6 & 80.2 & 81.9 & 78.7 & 69.9\\
PPTr+C2P & 150  & 81.1 & 84.0 & 85.4 & 82.5 & 74.1\\
PPTr+ours & 150  & \textbf{83.7} & \textbf{86.0} & \textbf{87.3} & \textbf{83.2} & \textbf{76.0}\\
\noalign{\smallskip}
\hline
\end{tabularx}
}
\vspace{-4mm}
\end{center}
}
\end{table}

\subsection{Fine-tuning on HOI4D semantic segmentation}
\label{sec:HOI4Dss}

\textbf{Setup.} To verify that our approach can also be effective on fine-grained tasks, we conducted further experiments on HOI4D for 4D semantic segmentation. For each point cloud frame, there are 8192 points. We follow the official data split of HOI4D with 2971 training scenes and 892 test scenes. 
During representation learning and training/fine-tuning, we randomly select 1/5 of the whole data to form one epoch for efficient training. We use mean IoU(mIoU) \% as the evaluation metric and 39 category labels are used to calculate it.
Considering our representation learning method prefers long sequence which has relatively abundant temporal information while the limitation of GPU memory, we set the sequence length as 10 and num points per frame as 4096. Fine-tuning and testing are performed on sequence length of 3 to be consistent with the baseline.

\textbf{Result.} As reported in Tab.~\ref{table:hoi4dss}, there is still a performance improvement on the 4D semantic segmentation task, which also shows the effectiveness of our approach for fine-grained feature understanding.
Compared with previous methods, our method can effectively extract features with high representational capabilities by introducing cross-modal learning.

\begin{table}
\setlength{\tabcolsep}{0.15mm}
\scriptsize{
\begin{center}
\caption{Semantic segmentation on HOI4D dataset}
\label{table:hoi4dss}
\newcolumntype{Y}{>{\centering\arraybackslash}X}
{
\begin{tabularx}{\columnwidth}{>{\centering} m{0.47\columnwidth}|>{\centering} m{0.2\columnwidth}|Y}
\hline\noalign{\smallskip}
Method & Frames   & mIoU\\
\noalign{\smallskip}
\hline
\noalign{\smallskip}

P4Transformer & 3	 & 40.1 \\
P4Transformer+C2P & 3 & 41.4 \\
P4Transformer+ours & 3 & \textbf{42.1} \\
\noalign{\smallskip}
\hline
\noalign{\smallskip}
PPTr & 3 & 41.0 \\
PPTr+STRL & 3  & 41.2 \\
PPTr+VideoMAE & 3 & 41.3 \\
PPTr+C2P & 3  & 42.3 \\
PPTr+ours & 3  & \textbf{43.8} \\
\noalign{\smallskip}
\hline
\end{tabularx}
}
\vspace{-2mm}
\end{center}
}
\end{table}

\subsection{Finetune image video encoder on action segmentation}
\label{sec:2d}
\textbf{Setup.} To demonstrate the representational capabilities of the image video encoder, we conduct experiments on HOI4D action segmentation using image video as input. 
We consider three representative methods as baseline backbone: MS-TCN\cite{farha2019ms}, MS-TCN++\cite{li2020ms} and Asformer\cite{yi2021asformer}. Followed by these work, we also use the I3D\cite{carreira2017quo} feature to train the network and use the pre-trained Transformer as initialization. We use the temporal resolution at 15 fps, and the dimension of the I3D feature for each frame is 2048-d. The following three metrics are reported: framewise accuracy (Acc), segmental edit distance, as well as segmental F1 scores at the overlapping thresholds of 10\%, 25\%, and 50\%. Overlapping thresholds are determined by the IoU ratio.

\begin{table}
\setlength{\tabcolsep}{0.15mm}
\scriptsize{
\begin{center}
\caption{3D Action segmentation on HOI4D dataset}
\label{table:hoi4d2das}
\newcolumntype{Y}{>{\centering\arraybackslash}X}
{
\begin{tabularx}{\columnwidth}{>{\centering} m{0.3\columnwidth}|>{\centering} m{0.12\columnwidth}|>{\centering} m{0.125\columnwidth}|Y|Y|Y|Y}
\hline\noalign{\smallskip}
Method & Frames   & Acc & Edit & F1@10 & F1@25 & F1@50\\
\noalign{\smallskip}
\hline
\noalign{\smallskip}
MS-TCN&150&44.2 &74.7 &55.6 &47.8&31.8\\
MS-TCN++ &150& 42.2&75.8 &54.7&46.5 &30.3\\
Asformer&150&46.8 &80.3 &58.9 &51.3&35.0\\
Asformer + Ours &150& \textbf{48.8} & \textbf{81.3} &\textbf{60.2} &\textbf{54.8}&\textbf{37.9}\\
\noalign{\smallskip}
\hline
\end{tabularx}
}
\vspace{-2mm}
\end{center}
}
\end{table}

\textbf{Result.} The results in Tab.~\ref{table:hoi4d2das} indicate that our method achieves powerful representation capabilities through image video encoder pre-training. It is worth noting that the action segmentation performance on image videos is significantly lower than that of point cloud videos in Table~\ref{table:hoi4das}. This may be due to the accurate three-dimensional coordinate information, point cloud video can provide the spatial position, shape, and motion trajectory of objects which gives point cloud videos an advantage in motion analysis.

\subsection{Ablations and Analysis}
\label{sec:ablation}
In this section, we first conduct an ablation study to verify the design of our method. we conduct all experiments on HOI4D 4D action segmentation task. We also conduct experiments to show whether image video and point cloud video can benefit each other when both modalities are available. We also evaluate our method under limited training data. 
 
\subsubsection{Ablation study}
Our core idea is that point cloud video and image video can complement each other, so we introduce cross-modal contrastive learning to facilitate communication between the two modalities for obtaining powerful representations. To verify our idea, we do ablation studies to show the effectiveness of each objective. We can find the accuracy drops 3.6\% without cross-modal contrastive loss and 2.0\%
without intra-modal contrastive learning. 
Besides, we can find the accuracy drops 1.4\% without frame level contrastive learning and 2.8\%
without video-level contrastive learning. 
The above results demonstrate the effectiveness of each objective.

\subsubsection{Multi-modal Fusion v.s Cross-modal pre-training}
We conducted experiments to verify whether the fusion of two modalities can also improve performance. As shown in the Tab.~\ref{table:hoi4dasfuse}, when using a single modality, point cloud videos achieve the best performance. When we cascade the features of the two modalities, noticeable performance improvement can be observed, but the fusion result is still lower than the performance that pre-trained point cloud video networks can achieve. This indicates that our pre-training method significantly promotes communication between cross-modalities and provides a good pre-trained model for point cloud video encoders. Additionally, it is worth mentioning that in real testing scenarios, it may not be possible to obtain data from both modalities simultaneously, which also hinders the possibility of feature fusion.

\begin{table}
\setlength{\tabcolsep}{0.15mm}
\scriptsize{

\begin{center}
\caption{Action segmentation on HOI4D dataset}
\label{table:hoi4dasfuse}
\newcolumntype{Y}{>{\centering\arraybackslash}X}
{
\begin{tabularx}{\columnwidth}{>{\centering} m{0.3\columnwidth}|>{\centering} m{0.12\columnwidth}|>{\centering} m{0.125\columnwidth}|Y|Y|Y|Y}
\hline\noalign{\smallskip}
Method & Frames   & Acc & Edit & F1@10 & F1@25 & F1@50\\
\noalign{\smallskip}
\hline
\noalign{\smallskip}
Asformer&150& 46.8 &80.3 &58.9 &51.3&35.0\\
Asformer + Ours &150& 48.8 & 81.3 &60.2 &54.8&37.9\\
\hline
PPTr & 150 &  77.4 & 80.1 & 81.7 & 78.5 & 69.5\\
PPTr+ours & 150  & \textbf{83.7} & \textbf{86.0} & \textbf{87.3} & \textbf{83.2} & \textbf{76.0}\\
\hline
PPTr + Asformer&150& 82.7 &84.7& 85.2 &81.9 &74.3\\
\noalign{\smallskip}
\hline
\end{tabularx}
}
\vspace{-2mm}
\end{center}
}
\end{table}

\begin{table}[t]
\setlength{\tabcolsep}{0.15mm}
\scriptsize{
\begin{center}
\caption{Data-efficient learning on HOI4D Action Segmentation.}
\label{table:hoi4das_efficient}
\newcolumntype{Y}{>{\centering\arraybackslash}X}
{
\begin{tabularx}
{\columnwidth}{>{\centering} m{0.15\columnwidth}|>{\centering} m{0.2\columnwidth}|>{\centering} m{0.3\columnwidth}|Y}
\hline\noalign{\smallskip}
\%Data & Scratch   & C2P & Ours\\
\noalign{\smallskip}
\hline
\noalign{\smallskip}

10\% & 44.0  & 53.4 & 57.8\\
20\% & 53.9 & 69.9 & 74.1\\
40\% & 69.9 & 75.4 & 76.9\\
80\% & 76.7 & 79.0 & 80.2\\
\noalign{\smallskip}
\hline
\end{tabularx}
}
\vspace{-2mm}
\end{center}
}
\end{table}

\subsubsection{Data-efficient 4D Representation Learning}
\label{sec:data-eff}

We evaluate our method under limited training data on the HOI4D Action Segmentation task. For all data-efficient experiments, our limited data are randomly sampled from the full dataset of HOI4D Action Segmentation dataset.  As shown in Tab.~\ref{table:hoi4das_efficient}, our pre-training method shows consistently outstanding performance in the case of lack of data, compared with previous methods. This indicates that our method can still formulate a strong 4D representation under limited data.

%% file: tex/conclusion.tex
This paper introduces CrossVideo, which applies self-supervised cross-modal contrastive learning in point cloud video understanding. We propose to leverage the cross-modal relationship between point cloud videos and image videos to learn meaningful feature representations. We employ intra-modality and cross-modality contrastive learning to facilitate effective point cloud video understanding. Additionally, we propose to contrast features of the two modalities at different levels. Experimental results demonstrate that our method outperforms previous state-of-the-art approaches.